\documentclass[final]{cvpr}

\usepackage{times}
\usepackage{epsfig}
\usepackage{graphicx}
\usepackage{amsmath}
\usepackage{amssymb}

\usepackage{graphicx}
\usepackage{bm}
\usepackage{subcaption}
\usepackage{multirow, eucal}
\usepackage[flushleft]{threeparttable}
\usepackage{booktabs}
\usepackage{algorithm}
\usepackage{algpseudocode}
\usepackage[labelfont=bf]{caption}
\usepackage{setspace}
\usepackage{float}
\usepackage{dsfont}
\usepackage{cite}
\usepackage{tabu}
\usepackage{booktabs}
\usepackage{soul}
\usepackage{arydshln}
\usepackage{algorithm}
\usepackage{algpseudocode}
\usepackage{subdepth}
\usepackage{textcomp}

\newcommand{\sgn}{\text{sgn}}

\usepackage[pagebackref=true,breaklinks=true,colorlinks,bookmarks=false]{hyperref}

\def\cvprPaperID{****} % *** Enter the CVPR Paper ID here
\def\confYear{CVPR 2021}
\begin{document}
%%%%%%%%% TITLE
\title{Learnable Companding Quantization for Accurate Low-bit Neural Networks}
\author{Kohei Yamamoto\\
Oki Electric Industry Co., Ltd.\\
{\tt\small yamamoto833@oki.com}
}
\maketitle
%%%%%%%%% ABSTRACT
\begin{abstract}
  Quantizing deep neural networks is an effective method for reducing memory consumption and improving inference speed, and is thus useful for implementation in resource-constrained devices.
	However, it is still hard for extremely low-bit models to achieve accuracy comparable with that of full-precision models.
	To address this issue, we propose learnable companding quantization (LCQ) as a novel non-uniform quantization method for 2-, 3-, and 4-bit models.
	LCQ jointly optimizes model weights and learnable companding functions that can flexibly and non-uniformly control the quantization levels of weights and activations.
	We also present a new weight normalization technique that allows more stable training for quantization.
	Experimental results show that LCQ outperforms conventional state-of-the-art methods and narrows the gap between quantized and full-precision models for image classification and object detection tasks.
	Notably, the 2-bit ResNet-50 model on ImageNet achieves top-1 accuracy of 75.1\% and reduces the gap to 1.7\%, allowing LCQ to further exploit the potential of non-uniform quantization.
\end{abstract}
%
%%%%%%%%% BODY TEXT
\vspace{-10pt}
\section{Introduction} \label{sec:intro}
Deep neural networks (DNNs) have been successfully applied to image-based tasks such as image classification and object detection, but their implementation in resource-constrained mobile or edge devices remains difficult, owing to the large number of required multiply--accumulate (MAC) operations and parameters.
To mitigate this problem, various techniques for compressing DNNs while maintaining performance have been proposed, such as pruning~\cite{He2020LearningFP}, knowledge distillation~\cite{Li2020FewSK}, low-rank approximation~\cite{Idelbayev2020LowRankCO}, and network quantization~\cite{Li2020AdditivePQ}.
Among these, network quantization is important as a way to effectively improve both memory consumption and inference speed.
However, network quantization is known to degrade performance of the original model in proportion to the amount of bit-width reduction.
\begin{figure}[t]
  \begin{center}
    \includegraphics[width=0.46\textwidth]{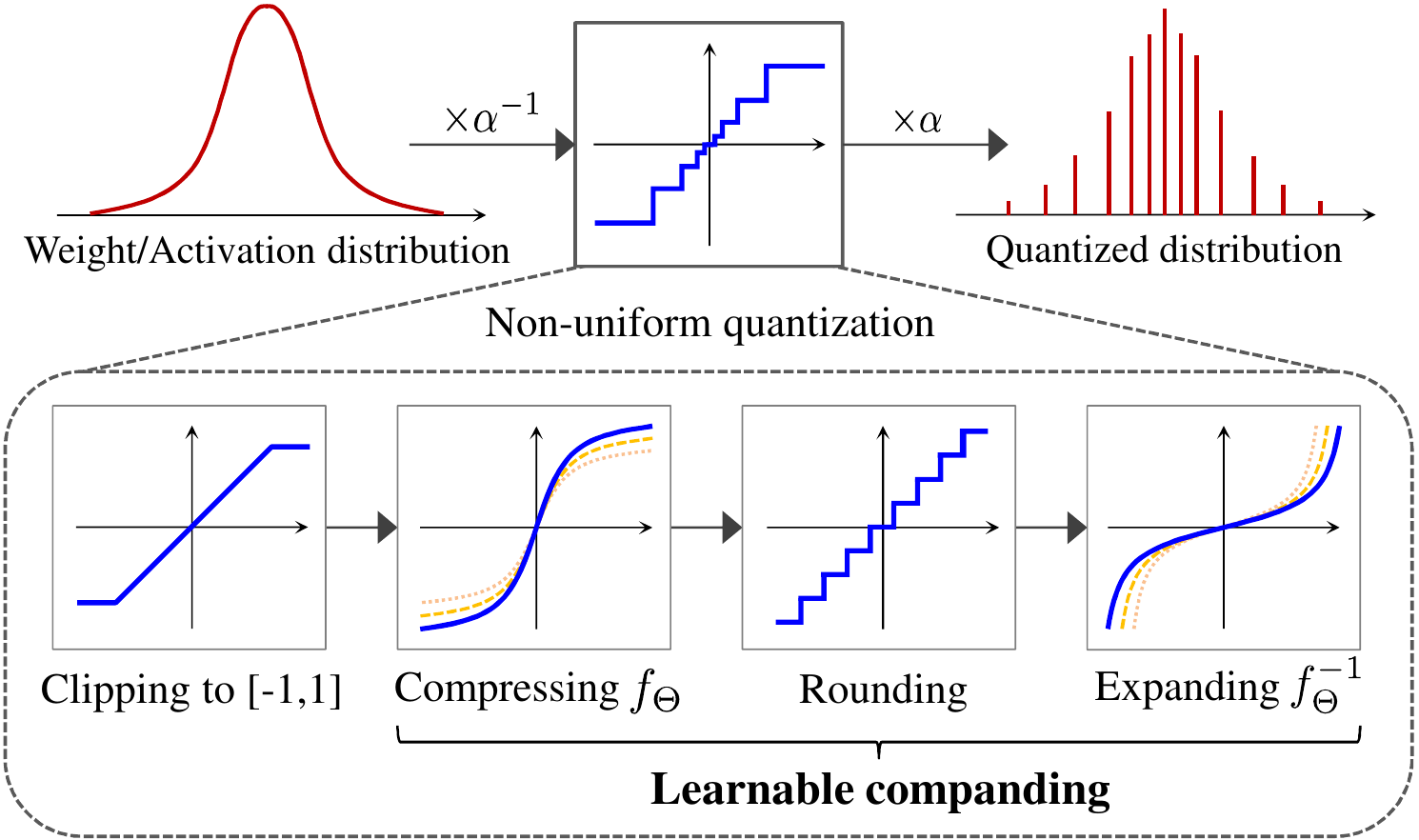}
    \caption{
    An overview of the proposed method.
  	Our non-uniform quantizer quantizes weights or activations with four functions, those for clipping, compressing, rounding, and expanding.
  	In particular, a composite function consisting of those except for the clipping function is generally called the companding function.
  	We formulate the companding function in a learnable form with a set of parameters $\Theta$ and jointly optimize it with clipping parameter $\alpha$ and the other parameters in the model.
    }
    \label{fig:overview}
    \vspace{-18pt}
  \end{center}
\end{figure}
\par In network quantization, the weights or activations of DNNs are typically discretized by a quantization function.
Although the quantization function is not differentiable, a straight-through estimator (STE)~\cite{Bengio2013EstimatingOP} can be used to approximate the gradient calculation for backpropagation.
Quantization functions are divided into two types: uniform and non-uniform quantization, in which input values are respectively linearly and nonlinearly discretized.
Because the weight or activation distribution is empirically dissimilar to the uniform distribution, non-uniform quantization can be expected to further reduce quantization and prediction errors than can uniform quantization via proper optimization.
For example, previous works on non-uniform quantization have attempted to use fixed and logarithmic quantization levels~\cite{Miyashita2016ConvolutionalNN, Li2020AdditivePQ} or learnable quantization levels that minimize quantization errors~\cite{Zhang_2018_ECCV}.
\par However, it is not easy to estimate effective quantization levels accurately, especially in low-bit models, where accuracy is often inferior to that of uniform quantization methods.
This paper thus aims to exploit the potential of non-uniform quantizers and further bridge the accuracy gap between quantized and full-precision models.
\par We propose a non-uniform quantization method called learnable companding quantization (LCQ). Figure~1 shows an overview of LCQ.
Our method is based on a companding (compressing and expanding) technique~\cite{compandor} that is widely used in telecommunication and signal processing to reduce the bit-width of input signals by nonlinearly limiting their dynamic range.
We assume that companding is effective for quantization of DNNs in two aspects.
The first is that the scale remains unchanged between inputs and outputs by using a nonlinear function and its inverse function, and that maintaining scale reduces quantization error and stabilizes training via backpropagation.
The second is that if the companding function is differentiable, its parameters can be optimized to directly minimize task loss.
Then, since the parameters are updated with a sum of the two gradients from the paths of before and after rounding, they can be trained with a large quantization influence.
Specifically, we formulate a learnable piecewise linear function as a nonlinear compressing function, allowing flexible and non-uniform control of quantization levels by optimization.
\par We also propose a new weight normalization method that improves accuracy by restoring the standard deviation of quantized weights to its original level, and we discuss a training trick for efficient inference with lookup tables.
\par Our main contributions are summarized as follows:
\par
\vspace{-5pt}
\begin{itemize}
  \setlength\itemsep{-3pt}
  \item
  We propose a LCQ method as a novel non-uniform quantization method, which optimizes non-uniform quantization levels to minimize task loss by training the learnable companding function.
  \item
  We present a simple normalization method for weight quantizers called limited weight normalization (LWN) that results in stable training and better solutions.
  \item
  We evaluate LCQ on various networks for image classification and object detection tasks, and the results show promising performance on the CIFAR-10/100, ImageNet, and COCO datasets.
\end{itemize}
\section{Related Works} \label{sec:related}
To quantize weights and activations in training, two operations are often applied in sequence: ``clipping'' and ``quantizing'' of inputs.
Quantization errors occur in each process, so various methods have been proposed to reduce them.\footnote{Relations to LCQ are also explained in the supplementary material~\ref{sec:supp_relation}.}
\par{\bf Clipping technique.}
To perform quantization, clipping is first applied to constrain the value range of inputs.
The simplest way to determine the clipping threshold is to use a given fixed value, but doing so does not adapt to variations in the dynamic range of the input values during training.
To address this issue, Jacob \etal~\cite{Jacob2018QuantizationAT} proposed a method that uses as the threshold the maximum value of the input tracked by the exponential moving averaging.
Choi \etal~\cite{Choi2018PACTPC} proposed a method that treats the threshold as a learnable parameter, optimizing it to minimize the task loss along with the weights.
Zhao \etal~\cite{Zhao2020Linear} used a simulated gradient to estimate the near-optimal threshold in every iteration.
Several prior works~\cite{Li2020AdditivePQ, Esser2020LearnedSS, Uhlich2020MixedPD} have proposed improved formulations of the learnable threshold approach, updating the parameter with a gradient calculated from residuals between the pre-quantized and quantized values.
\par{\bf Uniform quantization.}
Uniform quantization maps a clipped value to one of equally spaced discrete levels.
Although such mapping is performed with a nondifferentiable step function, STE~\cite{Bengio2013EstimatingOP} is often applied to approximate the gradient calculation and to enable parameter updates based on backpropagation.
Gong \etal~\cite{Gong2019DifferentiableSQ} proposed a method for mitigating gradient approximation errors incurred by using STE, representing the quantization function as a concatenation of several tanh functions and training their shape parameter to gradually converge on the step function.
Li \etal~\cite{Li_2019_CVPR} applied uniform quantization to object detection models with batch normalization folding.
Zhuang \etal~\cite{Zhuang_2018_CVPR} proposed a progressive two-stage quantization approach.
Jung \etal~\cite{Jung2019LearningTQ} introduced parameterized quantization intervals and optimized them to minimize task loss.
Liu \etal~\cite{Liu2019LearningLN} used a scheme that does not apply STE, instead using a weighted average of pre- and post-quantization values to gradually shift to the quantized values.
Zhuang \etal~\cite{Zhuang_2020_CVPR} proposed a training scheme using an auxiliary module connected to a low-bit network, providing it with the gradient from other loss.
\par{\bf Non-uniform quantization.}
Since DNN weights and activities are empirically non-uniformly distributed, non-uniform quantization, which discretizes inputs into unequal levels, should work effectively.
Han \etal~\cite{han2015deep} uses $k$-means clustering as a method of quantization to share weights.
Xu \etal~\cite{XuAAAI1816479} applies the same clustering strategy, but by gradually sharing weights in a layer-by-layer manner.
Miyashita \etal~\cite{Miyashita2016ConvolutionalNN} introduced non-uniform quantization using a powers-of-two (PoT) function and showed that multiplication in DNNs can be replaced by cheaper bit-shift operations.
Polino \etal~\cite{polino2018model} formulated quantization levels as learnable parameters and trained them with gradient descent and distillation methods.
Zhang \etal~\cite{Zhang_2018_ECCV} proposed parameterized bases for quantization levels and sequentially estimated an analytical solution that minimizes quantization error.
Li \etal~\cite{Li2020AdditivePQ} proposed an additive PoT quantizer to solve the problem of PoT functions that map extremely low quantization levels to larger input values.
\begin{figure*}[t]
  \setlength{\belowcaptionskip}{-5pt}
    \centering
    \begin{subfigure}[b]{0.49\linewidth}
       \setlength{\abovecaptionskip}{-15pt}
        \centering
        \includegraphics[width=\linewidth]{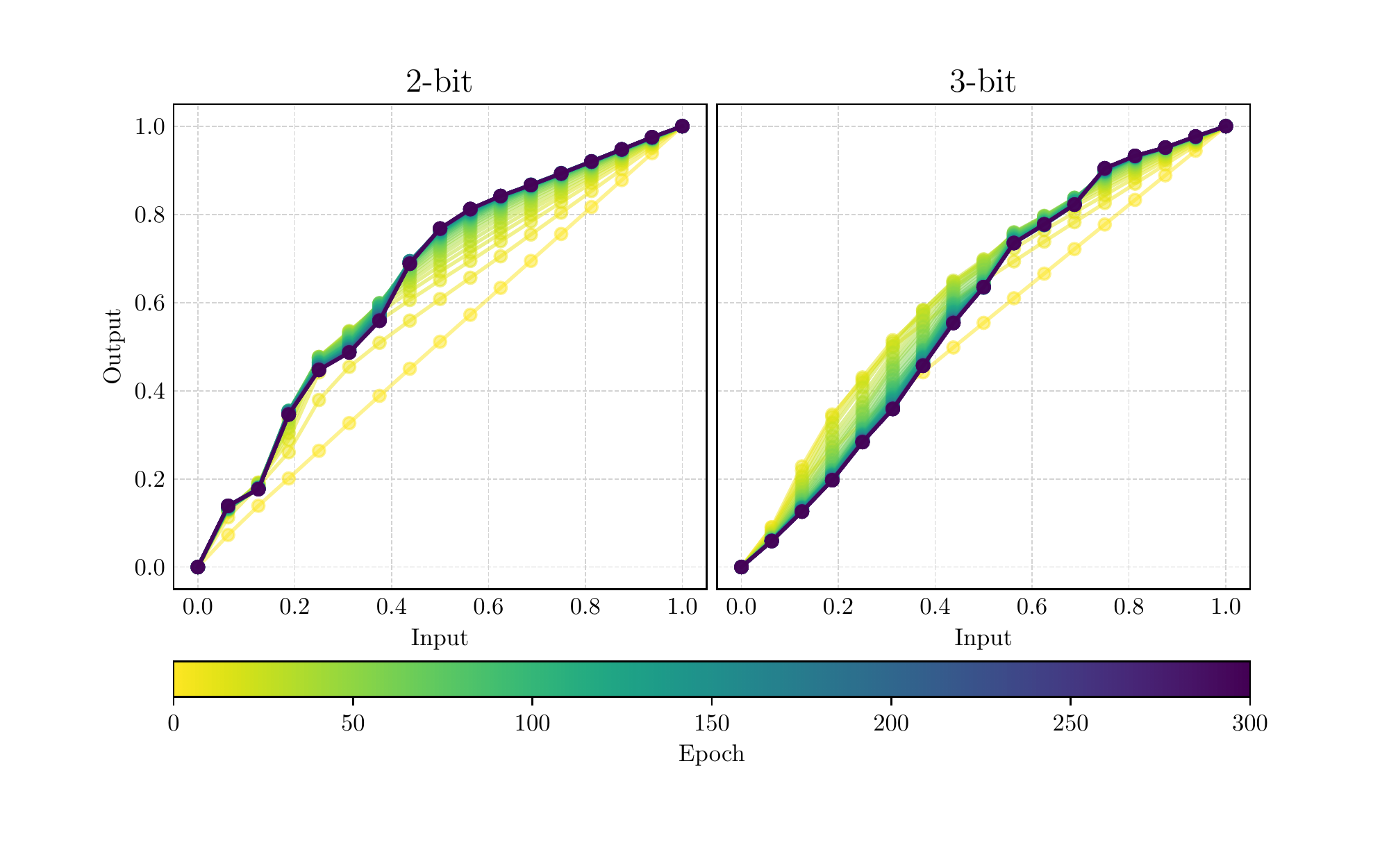}
        \caption{Compressing function $f_\Theta$}
        \label{fig:compressing}
    \end{subfigure}
    \begin{subfigure}[b]{0.49\linewidth}
        \setlength{\abovecaptionskip}{-15pt}
        \centering
        \includegraphics[width=\linewidth]{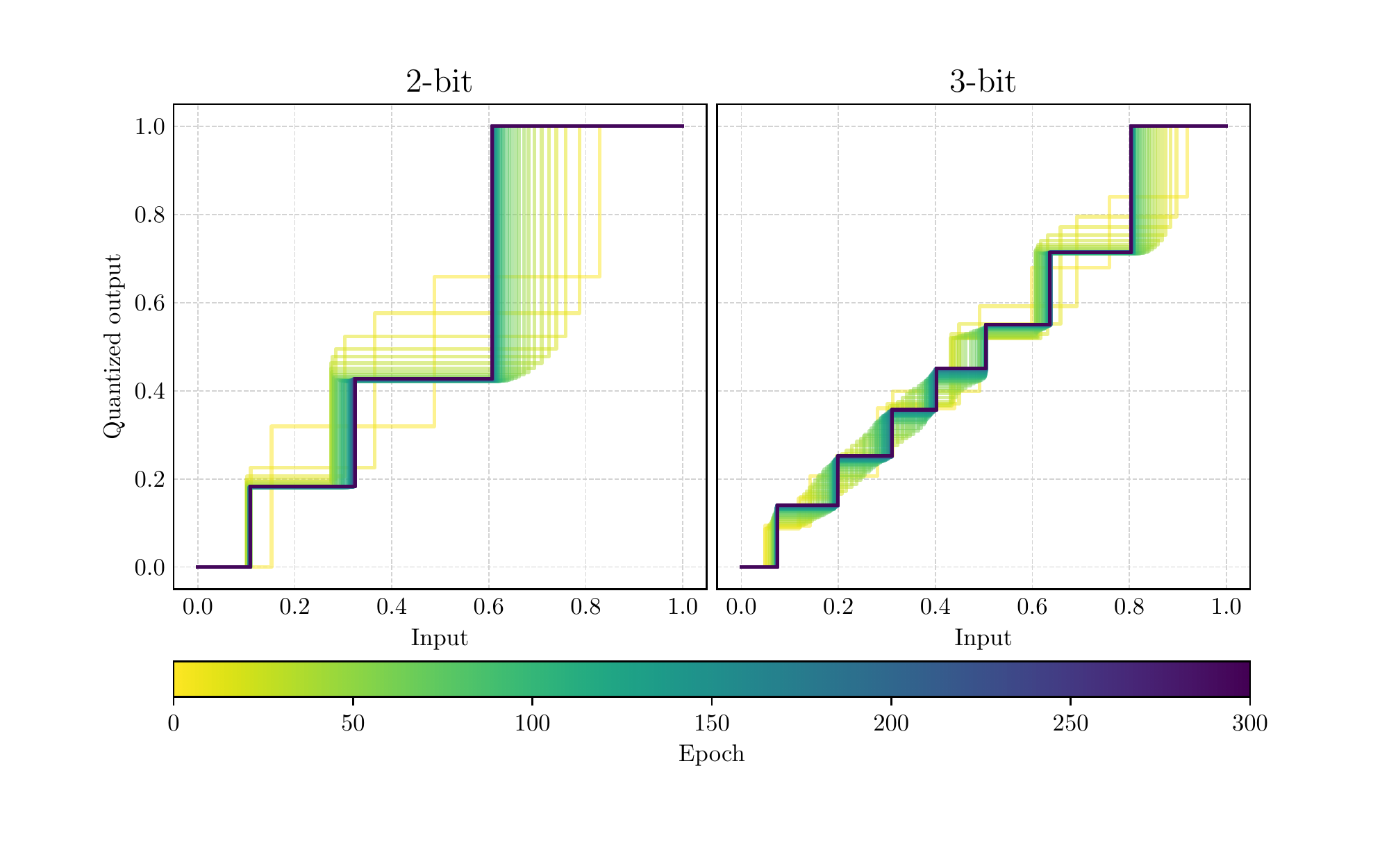}
        \caption{Companding function $g$}
        \label{fig:companding}
    \end{subfigure}
    \caption{
    Examples of the evolutionary process for the compressing ({\it left}) and companding ({\it right}) functions via training the LCQ quantizer for the ResNet-20 model on the CIFAR-10 dataset.
    Note that the 2- and 3-bit results of the companding function are generated using the compressing function with a corresponding number of bits.
    }
    \label{fig:evolution}
\end{figure*}
\section{Method}\label{sec:method}
In this section, we first provide a brief background of network quantization.
We then discuss details of the proposed method, including formulation of the LCQ quantizer, the LWN method, and a training trick for efficient inference.
\subsection{Preliminaries}\label{sec:preliminaries}
The goal of network quantization is to replace floating-point weights and activations for DNNs with lower bit-width ones to reduce memory consumption and speed up MAC operations.
For uniform quantization, a standard uniform quantizer $Q_U(x; \alpha)$ quantizes an input $x \in \mathbb{R}$ as
\begin{align}\label{eq:uniform_clip}
  Q_U(x; \alpha) = \sgn(x) \cdot \alpha
  \begin{cases}
   q_b \left(\frac{|x|}{\alpha}\right), & |x| < \alpha, \\
   1, & |x| \geq \alpha,
  \end{cases}
\end{align}
where $\alpha \in \mathbb{R}_{>0}$ is a clipping parameter, $q_b(\cdot)$ is a uniform quantization function, and the subscript $b \in \mathbb{N}$ is the bit-width.
This clipping operation reduces quantization error by multiplying the quantized value by $\alpha$ again to return it to its original value range.
Letting the clipped input be $v \in [0, 1)$, $q_b(v)$ can be represented as
\begin{align}
  q_b(v) = \frac{\left \lfloor{s \cdot v}\right \rceil}{s} ,
\end{align}
where the scaling factor $s$ becomes $s=2^{b-1}-1$ in the case of signed quantization or $s=2^b - 1$ in the case of unsigned quantization, and $\lfloor \cdot \rceil$ is a rounding function.
The quantization function $q_b(v)$ is not differentiable, because it contains a rounding function, but can be relaxed by STE~\cite{Bengio2013EstimatingOP} as $\partial q_b(v) / \partial v = 1$.
Similarly, the gradient for input $x$ does not vanish through the quantizer due to $\partial Q_U(x; \alpha) / \partial x = 1$ for $|x| < \alpha$.
When applying this quantizer to convolutional neural networks (CNNs), the weights and activations are independently quantized, just before the convolutional operations.
Then MAC operations in the convolution can be transformed to execute with integer precision, which can speed up the inference process~\cite{Jacob2018QuantizationAT}.
\subsection{Learnable Companding Quantization}\label{sec:lcq}
As Fig.~\ref{fig:overview} shows, our proposed LCQ is a non-uniform quantization method, without clipping, which mainly consists of three functions: a compressing function $f_{\Theta}(\cdot)$, a uniform quantization function $q_b(\cdot)$, and an expanding function $f_{\Theta}^{-1}(\cdot)$. A composite function formed from these three is generally called the companding function.
Note that $\Theta=\{\theta_1, \theta_2, \ldots, \theta_K\}$ is a common set of parameters for both functions.
Here, we denote the LCQ quantizer as
\begin{align}\label{eq:lcq}
  Q_{L}(x; \alpha, \Theta) = \sgn(x) \cdot \alpha
  \begin{cases}
   g\left(\frac{|x|}{\alpha}\right), & |x| < \alpha, \\
   1, & |x| \geq \alpha,
  \end{cases}
\end{align}
where
\begin{align}\label{eq:companding}
  g(v) = (f_{\Theta}^{-1} \circ q_b \circ f_{\Theta})(v)
\end{align}
is the companding function.
We then use a learnable piecewise linear function as the compressing function $f_\Theta(\cdot)$ and train its parameters $\Theta$ to minimize task loss.
The piecewise linear function is suited to fine-grained control of quantization levels, because it flexibly changes its slopes in proportion to the number of breakpoints (or intervals).
For example, Fig.~\ref{fig:evolution} shows the evolutionary processes of the piecewise linear function (Fig.~\ref{fig:compressing}) and its companding function (Fig.~\ref{fig:companding}) at the different bit-widths.
These figures show that the quantization levels and intervals of the companding function can be finely determined by changes in the slope of each interval of the piecewise linear function.
In this way, we generate accurate low-bit DNNs by giving the model the capability to directly tune the quantization levels.
\subsubsection{Detailed formulation} \label{sec:detailed_formulation}
Specifically, such a piecewise linear function needs to be monotonically increasing and to satisfy the constraint of an input range of $[0, 1)$ to account for the quantization function.
In our formulation, we first let the breakpoints that form the $k$-th interval (where $k \in \{ 1, 2, \ldots, K\}$) be equally spaced, meaning all interval lengths $\Delta = 1/K$.
We then prepare learnable parameters $\theta_k \in \Theta$ and use the softmax function to restrict their value range to $[0, 1]$, as in $\tilde{\theta}_k = \exp(\theta_k) / \sum_{i=1}^{K} \exp(\theta_i)$.
We further define the slope of the $k$-th interval as $\gamma_k = \tilde{\theta}_k / \Delta$, the total length of the $k$-th interval as $d_k = k\Delta$, and the cumulative sum of the output levels as $\beta_k = \sum_{i=1}^{k} \tilde{\theta}_i$, and we set $d_0 = 0$ and $\beta_0 = 0$.
With the above preparation, the piecewise linear compressing and expanding functions can be formulated as
\begin{align}
  f_{\Theta}(v) &= \sum_{k=1}^K \left( \gamma_k(v - d_{k-1}) + \beta_{k-1} \right) \mathds{1}_{[d_{k-1}, d_k)}(v), \label{eq:compress}
  \\
  f_{\Theta}^{-1}(v) &= \sum_{k=1}^K \left( \frac{v - \beta_{k-1}}{\gamma_k} + d_{k-1} \right) \mathds{1}_{[\beta_{k-1}, \beta_k)}(v), \label{eq:expand}
\end{align}
where $\mathds{1}_{\mathcal{C}}(v)$ is an indicator function that returns 1 if $v \in \mathcal{C}$ and 0 otherwise.
We finally use a gradient descent algorithm to optimize $\theta_k$ through $\gamma_k$ and $\beta_k$.
\subsubsection{Backpropagation for companding} \label{sec:backprop_companding}
By the chain rule, the gradient of our quantizer $Q_L(\cdot)$ with respect to $\theta_k$ can be written as
\begin{align} \label{eq:chain_rule}
  \frac{\partial Q_L}{\partial \theta_k} =
  \left(
  \frac{\partial Q_L}{\partial \gamma_k} \frac{\partial \gamma_k}{\partial \tilde{\theta}_k}
  +
  \frac{\partial Q_L}{\partial \beta_k} \frac{\partial \beta_k}{\partial \tilde{\theta}_k}
  \right)
  \frac{\partial \tilde{\theta}_k}{\partial \theta_k}.
\end{align}
Here, the gradients of $Q_L(\cdot)$ with respect to $\gamma_k$ and $\beta_k$ should be carefully calculated, because the compressing and expanding function may use intervals with different correspondences when going through the quantization function.
For simplicity, let $v_q$ be the output of the quantization and compressing function, i.e., $v_q = (q \circ f_\Theta)(v)$. Then the gradients of the companding function $g(v)$ can be represented as
\begin{align}
  \frac{\partial g(v)}{\partial \gamma_k}
  &\simeq
  \sum_{i=1}^K \left( \frac{v - d_{k-1}}{\gamma_i} I_{(k,i)}
  -
  \frac{v_q - \beta_{k-1}}{\gamma_k^2}I_{(i,k)} \right), \label{eq:deriv_gamma}
  \\
  \frac{\partial g(v)}{\partial \beta_k}
  &\simeq
  \sum_{i=1}^K \left( \frac{I_{(k,i)}}{\gamma_i}
  -
  \frac{I_{(i,k)}}{\gamma_k}  \right), \label{eq:deriv_beta}
\end{align}
where
\begin{align}
  I_{(i,j)} = \mathds{1}_{[d_{i-1}, d_i)}(v) \cdot \mathds{1}_{[\beta_{j-1}, \beta_j)}(v_q).
\end{align}
Note that we use the STE~\cite{Bengio2013EstimatingOP} approximation for the derivative of the quantization function, and that $v_q$ may exceed the upper bound on the value range $[0, 1)$ due to rounding, but in that case an inifinitesimal value $\varepsilon$ is subtracted from $v_q$ to keep it within the range.
The gradient of $Q_L(\cdot)$ with respect to $\gamma_k$ can then be written as
\begin{align} \label{eq:deriv_q_gamma}
  \frac{\partial Q_L}{\partial \gamma_k} \simeq
  \begin{cases}
    \sgn(x) \cdot \alpha \cdot \frac{\partial g \left( \frac{|x|}{\alpha} \right)}{\partial \gamma_k}, & |x| < \alpha,\\
    0, & |x| \geq \alpha.
  \end{cases}
\end{align}
Similarly, $\partial Q_L / \partial \beta_k$ is the replacement of $\gamma_k$ by $\beta_k$ in Eq.~(\ref{eq:deriv_q_gamma}).
Since the gradient contains the clipping parameter $\alpha$, the clipping effect is considered in the optimization of $\theta_k$.
\par The gradient of $g(\cdot)$ with respect to the clipped input $v$ is similarly affected by the quantization function as $\partial g(v) / \partial v = \sum_{i=1}^K \sum_{j=1}^K \gamma_i / \gamma_j \cdot I_{(i,j)}$.
However, since we have empirically found that the gradient may be too large when $\gamma_j$ is small and $\gamma_i$ is large, and that such a gradient makes the training unstable, we use $\partial g(v) / \partial v = 1$ instead.
Then the gradient of the quantizer $Q_L$ with respect to the input $x$ becomes $\partial Q_L / \partial x = 1$ for $|x| < \alpha$ and $0$ otherwise, like the uniform quantizer.
This strategy of not modifying (``straight-throughing'') the gradient for inputs has been used for other non-uniform quantization methods as well~\cite{Zhang_2018_ECCV, Li2020AdditivePQ}.
\subsubsection{Backpropagation for clipping} \label{sec:backprop_clip}
We estimate the clipping parameter $\alpha$ based on training, as in previous works~\cite{Li2020AdditivePQ, Esser2020LearnedSS, Uhlich2020MixedPD}.
We specifically update $\alpha$ based on the gradient of our quantizer $Q_L(\cdot)$, represented as
\begin{align}\label{eq:deriv_clip}
  \frac{\partial Q_L}{\partial \alpha} \simeq \sgn(x)
  \begin{cases}
    g \left( \frac{|x|}{\alpha} \right) - \frac{|x|}{\alpha}, & |x| < \alpha,\\
    1, & |x| \geq \alpha.
  \end{cases}
\end{align}
Note that $\alpha$ is jointly trained with companding parameters $\Theta$.
When parameterizing the compressing function, breakpoints related to the input interval are set to be equally spaced rather than trained, thereby preventing changes in the clipping parameter from having a significant effect on breakpoint locations, which would reduce the training efficiency.
Using equal spacing is less flexible, but this can be compensated for by increasing the number of breakpoints (or intervals).
\subsection{Limited Weight Normalization} \label{sec:lwn}
Li \etal~\cite{Li2020AdditivePQ} reported that clipping parameter training for each layer is stabilized when the weights are standardized with their mean and standard deviation before applying the quantizer.
There are two main reasons for this: because the weight distribution is zero-centered, satisfying the symmetry assumption in signed quantization, and because the clipping parameters are less sensitive to variations in the standard deviation.
\par However, considering that the quantized model is initialized with pretrained, full-precision weights to obtain good accuracy in many quantization methods~\cite{Li2020AdditivePQ, Esser2020LearnedSS, Jung2019LearningTQ}, weight normalization causes a gap for the output scale of the linear layer before and after quantization, which may negatively affect training.
Therefore, we also propose a method called limited weight normalization (LWN), which limits only the effective scope of normalization to the weight quantizer. LWN can be formulated as
\begin{align}
  \tilde{w} = \sigma_w \cdot Q_*\left( \frac{w - \mu_w}{\sigma_w} \right), \label{eq:lwn}
\end{align}
where $w \in \mathbb{R}$ is an input weight, $\mu_w$ and $\sigma_w$ are the sample mean and standard deviation of the weights, respectively, and $Q_*(\cdot)$ is any signed quantizer.
Note that the gradients for $\mu_w$ and $\sigma_w$ are not used to update all learnable parameters, as in ~\cite{Li2020AdditivePQ}.
The only difference between LWN and \cite{Li2020AdditivePQ} is whether the standard deviation is multiplied after quantization or not.
This simple method has the effect of not only restoring the standard deviation to its pre-normalized level in forward propagation, but also making the gradients for learnable quantizer parameters depend on the standard deviation in backpropagation.
We have empirically observed that this method is more stable and gives better solutions.
\subsection{Training for the LUT-based Inference} \label{sec:lut}
\begin{figure}[t]
  \begin{center}
    \includegraphics[width=0.42\textwidth]{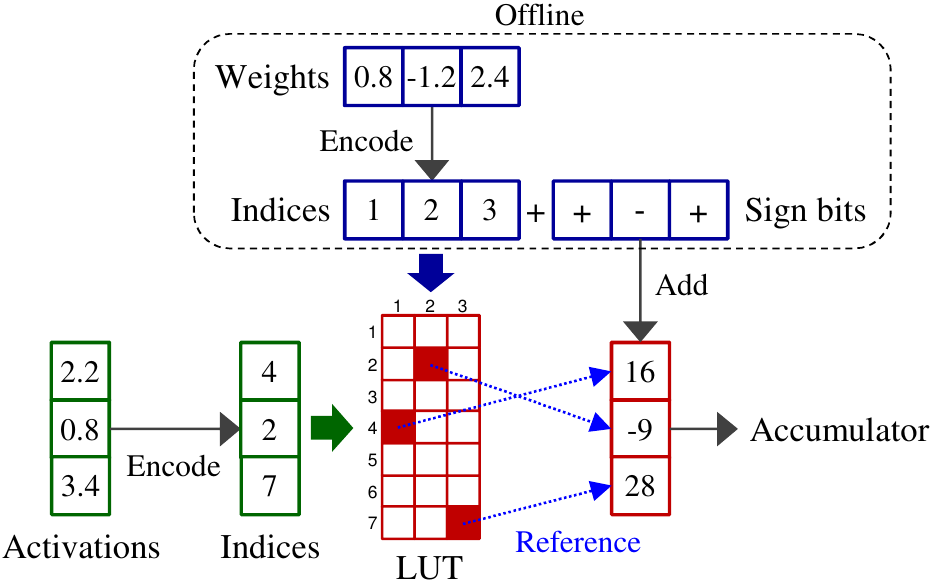}
    \caption{
    An example of the memory access with a LUT for inference (3-bit case).
    The weights can be pre-converted to the encoded low-bit indices.
    The LUT has precomputed multiplicative values for all output combinations generated by the weight and activation quantizers, and does not include zeros because they can simply be skipped.
    }
    \label{fig:lut}
    \vspace{-18pt}
  \end{center}
\end{figure}

In general, non-uniform quantization functions, including the companding function, output floating-point values, so the multiplication between weights and activations is also performed in floating-point, which is inefficient.
To speed up the operations during inference in deployment scenarios, it is often used to replace the multiplication with a memory access to a precomputed lookup table (LUT), as shown in Fig.~\ref{fig:lut}.
However, our method requires one LUT per convolutional or fully-connected layer, and thus incurs additional memory costs.
For example, with $b_w$ and $b_a$ for the number of bits in the signed weights and in the unsigned activations, the number of LUT elements $m$ becomes $m = (2^{b_w - 1} - 1)(2^{b_a} - 1)$.\footnote{Note that the sign bit is reduced because it can be applied afterwards, and the number of zeros is also reduced.}
Therefore, the additional memory cost per layer is simply $4m$ bytes for multiplication at the 32-bit floating-point precision.
The memory cost of LUTs should be reduced as much as possible, because this cost relates to the memory access speed and the accumulator capacity on dedicated devices such as field-programmable gate arrays (FPGAs).
\par To reduce the memory cost of LUTs, we apply the $b'$-bit uniform quantization function $q_{b'}(\cdot)$ immediately after the companding function $g(\cdot)$, where $b' \in \{t \in \mathbb{N} \mid t > b \}$ is the other bit-width (below, we call this the {\bf ``outer'' bit-width} for clarity).
Although this re-quantization introduces an extra quantization error, the effect on the accuracy is almost negligible if the outer bit-width is sufficiently larger than the original bit-width $b$, considering that quantization-aware training tends to quantize to around 8 bits with nearly no degradation in accuracy~\cite{Jacob2018QuantizationAT}.
Specifically, instead of Eq.~(\ref{eq:lcq}) for training, we use a slightly modified version of the quantizer $Q_L(\cdot)$,
\begin{align}\label{eq:lcq_prime}
  Q'_{L}(x; \alpha, \Theta) = \sgn(x) \cdot \alpha
  \begin{cases}
   (q_{b'} \circ g)\left(\frac{|x|}{\alpha}\right), & |x| < \alpha, \\
   1, & |x| \geq \alpha.
  \end{cases}
\end{align}
Note that since STE~\cite{Bengio2013EstimatingOP} is used for $q_{b'}(\cdot)$, all the backpropagation formulas in Sec.~\ref{sec:lcq} can also be applied to the quantizer $Q'_L(\cdot)$ as-is.\footnote{The training algorithm for a convolutional layer using $Q'_L(\cdot)$ is summarized in the supplementary material~\ref{sec:supp_algorithm}.}
Since all the scalar multiplications (e.g., by the clipping parameter $\alpha$, the scaling factor $s$ and the standard deviation $\sigma_w$ in LWN) after re-quantization can be moved after convolution at the inference time, the convolution can be performed with integer precision.
For the integer multiplication in the convolution, the bit-width of LUT elements is equal to the sum of the outer bit-width of activations and the one of weights.
Therefore, the more the outer bit-width is reduced, the more the bit-width of the LUT elements is reduced.
For example, letting the outer bit-widths for weights and activations be $b'_w$ and $b'_a$, as before, the memory cost of a LUT can be represented as $2^{-3}(b'_w + b'_a)m$ bytes.
The memory cost for $b'_w = b'_a = 8$ is two times smaller than the one for the 32-bit floating-point case.
The effect of changing the outer bit-width on accuracy is evaluated in the ablation study in Sec.~\ref{sec:ablation}.
\section{Experiments} \label{sec:exp}
This section evaluates the effectiveness of our method in comparison with conventional state-of-the-art uniform and non-uniform quantization methods using various models, such as ResNet~\cite{He2016DeepRL}, MobileNet-V2~\cite{Sandler2018MobileNetV2IR}, and RetinaNet~\cite{Lin_2017_ICCV}.\footnote{A discussion of the validity of the comparison between LCQ and uniform quantization methods is provided in the supplementary material \ref{sec:supp_validity}.}
We then report the results of ablation studies.
\par To evaluate our method, we used the CIFAR-10/100~\cite{Krizhevsky2009LearningML} and ImageNet (ILSVRC-2012)~\cite{Deng2009ImageNetAL} datasets for image classification tasks and the MS COCO~\cite{COCO} dataset for an object detection task.
The CIFAR-10/100 dataset contains 50k training images and 10k test images, with 10/100 classes.
The ImageNet dataset contains 1.2M training images and 50k test images, with 1,000 classes.
For the COCO dataset with 80 object classes, following~\cite{Lin_2017_ICCV, Zhuang_2020_CVPR} we used the {\it trainval35k} split (115k training images) and {\it minival} split (5k test images).
All experiments were implemented using PyTorch~\cite{NEURIPS2019_9015} and Cupy~\cite{cupy_learningsys2017}, and for the object detection task we also used Detectron2~\cite{wu2019detectron2}.
\subsection{Implimention Details} \label{sec:impl}
Unless otherwise specified, we used the following settings in all experiments.
\par {\bf Basic settings.}
We used signed and unsigned quantizers for weights and activations, respectively.
Note that a 2-bit signed quantization for weights implies ternarization. We instead applied the uniform quantizer $Q_U(\cdot)$ only for 2-bit weights (not for 2-bit activations), because ternarization eliminates the effect of companding.
Although quantization of the first and last layers significantly impacts accuracy, as in~\cite{Li2020AdditivePQ} we applied 8-bit quantization to both for further efficiency.
For both weights and activations, the number of intervals in the companding function and the outer bit-widths were set to $K = 16$ and $b' = 8$, respectively.
\par {\bf Optimization.}
We used the stochastic gradient descent algorithm with a Nesterov momentum of $0.9$ and cosine learning rate decay without restart~\cite{Loshchilov2017SGDRSG}.
All weights in the quantized model were initialized with pretrained weights at full precision, and we did not use the progressive initialization strategy~\cite{Zhuang_2018_CVPR, Jung2019LearningTQ}.
The clipping parameters for weights and activations were initialized as 3.0 and 8.0, respectively, and all companding parameters were initialized as 0; this is equivalent to the uniform quantization setting.
All gradients with respect to the clipping function for the weights were not zeroed out by applying STE~\cite{Bengio2013EstimatingOP}, as in~\cite{Li2020AdditivePQ}.
\par {\bf Architecture.}
Our unsigned activation quantizer can place a learnable upper bound on the inputs, but not a learnable lower bound.
However, a lower bound can be applied by adding a learnable bias term just before the unsigned quantization.
Implementationally, the bias term can be introduced by using batch normalization.
As in~\cite{Esser2020LearnedSS}, therefore, we used ResNet~\cite{He2016DeepRL} with pre-activation as a target architecture satisfying this condition.
We also used the same configuration as that of ResNet for the inverted residual blocks of MobileNet-V2~\cite{Sandler2018MobileNetV2IR}.
\subsection{Evaluation on CIFAR-10} \label{sec:cifar10}
We performed experiments for ResNet-20/56 on the CIFAR-10 dataset, training the quantized models over 300 epochs with an initial learning rate of $0.04$ for the weights and $0.02$ for the clipping and companding parameters, and with a mini-batch size of 128.
The weight decay was set to $10^{-4}$.
We adopted standard data augmentation techniques, namely random crop and horizontal flip.
\begin{table}[hbt!]
	\centering
	\caption{Top-1 accuracy (\%) for the 2/3/4-bit ResNet on the CIFAR-10 dataset.}
	\scalebox{0.9}{
	\begin{tabular}{c|l|ccc}
		Model &Method &W2/A2 &W3/A3 &W4/A4 \\\toprule
		\multirow{4}{*}{\shortstack[l]{ResNet-20\\~(FP: 93.4)}}
      &PACT~\cite{Choi2018PACTPC}    &89.7  &91.1  &91.7 \\
		  &LQ-Nets~\cite{Zhang_2018_ECCV} &90.2  &91.6  &-    \\
		  &APoT~\cite{Li2020AdditivePQ}    &91.0  &92.2  &92.3 \\
      &LCQ \small (Ours)     &{\bf91.8}  &{\bf92.8}  &{\bf93.2} \\\hline
    \multirow{2}{*}{\shortstack[l]{ResNet-56\\~(FP: 94.5)}}
      &APoT~\cite{Li2020AdditivePQ}    &92.9  &93.9  &94.0 \\
      &LCQ \small (Ours)     &{\bf93.5}  &{\bf94.6}  &{\bf94.7} \\\hline
	\end{tabular}}
	\label{tab:acc_cifar10}
\end{table}
\par Table~\ref{tab:acc_cifar10} compares the accuracy of the proposed and conventional methods at three bit-widths for the CIFAR-10 dataset.
In the table, for example, ``W2/A2'' indicates the case where the weights and activations are both quantized to 2 bits, ``FP'' indicates accuracy in the full precision case, and ``-'' indicates no reported result.
For ResNet-20, our LCQ shows better performance than do the uniform quantization method PACT~\cite{Choi2018PACTPC} and the non-uniform quantization methods LQ-Nets~\cite{Zhang_2018_ECCV} and APoT~\cite{Li2020AdditivePQ} at all bit-widths from 2 to 4.
Figure~\ref{fig:companding} shows examples of trained quantization levels for the 2/3-bit ResNet-20.
As shown in the 3-bit case in the figure, quantization levels for an input of around 0.4 are relatively dense and indicate an important value range for loss reduction.
The LCQ results were also better for ResNet-56 than was the APoT method.
Although APoT has fine-grained quantization levels due to the powers-of-two combination, unlike LCQ, the levels are not learnable.
\subsection{Evaluation on ImageNet} \label{sec:imagenet}
We evaluated the performance of LCQ for ResNet-18/34/50 and MobileNet-V2 on the ImageNet dataset.
With an initial learning rate of $0.1$ for the weights and an initial learning rate of $0.01$ for the clipping and companding parameters, the models were trained over 120 epochs with a mini-batch size of 1024 for ResNet-18/34 and 512 for both ResNet-50 and MobileNet-V2.
In addition, we applied a warm-up method~\cite{Goyal2017AccurateLM} for the first 5 epochs and increased the learning rate linearly from $10^{-4}$ to the initial value.
The weight decay was set to $4 \times 10^{-5}$.
The training images were resized, cropped to $224 \times 224$ pixels and randomly flipped horizontally.
The test images were center-cropped to $224 \times 224$ pixels.
\begin{table}[hbt!]
	\centering
	\caption{Top-1 accuracy (\%) for the 2/3/4-bit ResNet on the ImageNet dataset.}
	\scalebox{0.9}{
	\begin{tabular}{c|l|ccc}
		Model &Method &W2/A2 &W3/A3 &W4/A4 \\\toprule
		\multirow{6}{*}{\shortstack[l]{ResNet-18\\~(FP: 70.4)}}
      &LQ-Nets~\cite{Zhang_2018_ECCV}      &64.9  &68.2  &69.3 \\
		  &DSQ~\cite{Gong2019DifferentiableSQ} &65.2  &68.7  &69.6 \\
		  &QIL~\cite{Jung2019LearningTQ}       &65.7  &69.2  &70.1 \\
      &APoT~\cite{Li2020AdditivePQ}        &67.3  &69.9  &70.7 \\
      &LSQ~\cite{Esser2020LearnedSS}       &67.6  &70.2  &71.1 \\
      &LCQ \small (Ours)     &{\bf68.9}  &{\bf70.6}  &{\bf71.5} \\\hline
    \multirow{6}{*}{\shortstack[l]{ResNet-34\\~(FP: 74.2)}}
      &LQ-Nets~\cite{Zhang_2018_ECCV}      &68.8  &71.9  &- \\
      &DSQ~\cite{Gong2019DifferentiableSQ} &70.0  &72.5  &72.8 \\
      &QIL~\cite{Jung2019LearningTQ}       &70.6  &73.1  &73.7 \\
      &APoT~\cite{Li2020AdditivePQ}        &70.9  &73.4  &73.8 \\
      &LSQ~\cite{Esser2020LearnedSS}       &71.6  &73.4  &74.1 \\
      &LCQ \small (Ours)     &{\bf72.7}  &{\bf74.0}  &{\bf74.3} \\\hline
    \multirow{6}{*}{\shortstack[l]{ResNet-50\\~(FP: 76.8)}}
      &PTG.~\cite{Zhuang_2018_CVPR}        &70.0  &-     &75.7 \\
      &LQ-Nets~\cite{Zhang_2018_ECCV}      &71.5  &74.2  &75.1 \\
      &APoT~\cite{Li2020AdditivePQ}        &73.4  &75.8  &76.6 \\
      &LSQ~\cite{Esser2020LearnedSS}       &73.7  &75.8  &{\bf76.7} \\
      &Auxi~\cite{Zhuang_2020_CVPR}        &73.8  &-     &- \\
      &LCQ \small (Ours)     &{\bf75.1}  &{\bf76.3}  &76.6 \\\hline
	\end{tabular}}
	\label{tab:acc_imagenet}
	\vspace{-0.8em}
\end{table}
\par Table~\ref{tab:acc_imagenet} compares quantization performance with state-of-the-art conventional methods.
For almost all models, LCQ outperformed the conventional uniform~\cite{Gong2019DifferentiableSQ,Jung2019LearningTQ,Esser2020LearnedSS,Zhuang_2020_CVPR,Zhuang_2018_CVPR} and non-uniform~\cite{Zhang_2018_ECCV,Li2020AdditivePQ} methods on the ImageNet validation set.
Accuracy improvements over the conventional methods at 2 bits were particularly remarkable, with a maximum improvement of $1.3\%$ points for both ResNet-18 and ResNet-50.
These results suggest that fine-tuning of the quantization levels by the companding function improved accuracy in relatively large datasets.
\begin{table}[hbt!]
	\centering
	\caption{Accuracy (\%) of the 4-bit MobileNet-V2 on the ImageNet dataset.}
	\scalebox{0.9}{
	\begin{tabular}{c|l|cc}
		Model &Method &Top-1 acc. &Top-5 acc. \\\toprule
    \multirow{3}{*}{\shortstack[l]{MobileNet-V2\\~~~~(FP: 71.9)}}
    &DSQ~\cite{Gong2019DifferentiableSQ}  &64.8  &- \\
    &LLSQ~\cite{Zhao2020Linear}           &67.4  &88.0 \\
    &LCQ \small (Ours)       &{\bf 70.8} &{\bf 89.7} \\\hline
	\end{tabular}}
	\label{tab:acc_mnetv2}
	\vspace{-0.8em}
\end{table}
\par As Table~\ref{tab:acc_mnetv2} shows, we observed that LCQ achieved relatively good accuracy even for a compact and efficient architecture, 4-bit MobileNet-V2 (W4/A4).
Due to the low redundancy, the accuracy difference from the full-precision model was more than $1\%$ point and was not as close as in the case of the ResNet models in Table~\ref{tab:acc_imagenet}.
\subsection{Evaluation on COCO}
We used RetinaNet~\cite{Lin_2017_ICCV} with ResNet as the backbone to evaluate the proposed method on the COCO dataset. An initial learning rate of 0.005 was used for the weights and 0.001 for both the clipping and companding parameters.
The weight decay was set to $10^{-4}$ and the warm-up method~\cite{Goyal2017AccurateLM}, which increases the learning rate linearly from 0 to the initial value, was used for first 1k iterations.
The batch size was set to 16 and the models were trained over 90k iterations.
We resized both training and test images to have 800 pixels on shorter edges,
randomly flipping the training images horizontally as data augmentation.
Following the observation in~\cite{Zhuang_2020_CVPR}, the prediction head of RetinaNet was not shared between features of different resolutions, except for the final layers for regression and classification.
In addition, we inserted the batch normalization just before all convolutional layers for both the feature pyramid network (FPN) and the prediction heads, and synchronously updated all batch statistics during training.
We did not quantize only the last layers.
All other settings were in accordance with the original settings in~\cite{Lin_2017_ICCV}.
\begin{table}[hbt!]
	\centering
	\caption{APs for the 4-bit RetinaNet on the COCO dataset.}
  \setlength{\tabcolsep}{4pt}
  \renewcommand{\arraystretch}{1.1}
	\scalebox{0.86}{
	\begin{tabular}{c|l|cccccc}
	  Backbone &Method  &AP &AP$_{50}$  &AP$_{75}$ &AP$_{S}$  &AP$_{M}$  &AP$_{L}$ \\ \toprule
    \multirow{5}{*}{ResNet-18}
    &FP                           &33.2 &52.3 &34.8 &18.7 &35.6 &43.7 \\
   	&FQN~\cite{Li_2019_CVPR}      &28.6 &46.9 &29.9 &14.9 &31.2 &38.7 \\
   	&Auxi~\cite{Zhuang_2020_CVPR} &31.9 &50.4 &33.7 &16.5 &34.6 &{\bf 42.3} \\
    &APoT~\cite{Li2020AdditivePQ} &32.4 &51.2 &34.0 &18.4 &34.6 &42.2 \\
    &LCQ \small (Ours) &{\bf 32.7} &{\bf 51.7} &{\bf 34.2} &{\bf 18.6} &{\bf 35.2} &{\bf 42.3} \\ \hline
    \multirow{4}{*}{ResNet-34}
    &FP                           &37.2 &57.0 &39.4 &21.4 &40.4 &48.9 \\
		&FQN~\cite{Li_2019_CVPR}      &31.3 &50.4 &33.3 &16.1 &34.4 &41.6 \\
		&Auxi~\cite{Zhuang_2020_CVPR} &34.7 &53.7 &36.9 &19.3 &38.0 &45.9 \\
    &LCQ \small (Ours) &{\bf 36.4} &{\bf 55.9} &{\bf 38.7} &{\bf 21.2} &{\bf 40.0} &{\bf 46.6} \\ \hline
    \multirow{4}{*}{ResNet-50}
    &FP                           &38.3 &58.3 &40.9 &21.5 &42.4 &49.5 \\
    &FQN~\cite{Li_2019_CVPR}      &32.5 &51.5 &34.7 &17.3 &35.6 &42.6 \\
		&Auxi~\cite{Zhuang_2020_CVPR} &36.1 &55.8 &38.9 &{\bf 21.2} &39.9 &46.3 \\
    &LCQ \small (Ours) &{\bf 37.1} &{\bf 57.0} &{\bf 39.6} &{\bf 21.2} &{\bf 40.8} &{\bf 47.1}\\ \hline
	\end{tabular}}
	\label{tab:ap_coco_4bit}
  \vspace{-0.8em}
\end{table}
\par Table~\ref{tab:ap_coco_4bit} compares the COCO average precision (AP) metrics for the 4-bit RetinaNet (W4/A4) with the different backbones.
``FP'' indicates the full precision case.
LCQ showed more favorable results than did the conventional methods for all the 4-bit models, especially for ResNet-34, which differed by $1.7\%$ points in AP over Auxi~\cite{Zhuang_2020_CVPR} with the uniform quantization.
\begin{table}[hbt!]
	\centering
	\caption{APs for the 3-bit RetinaNet on the COCO dataset.}
  \setlength{\tabcolsep}{4pt}
  \renewcommand{\arraystretch}{1.1}
	\scalebox{0.86}{
	\begin{tabular}{c|l|cccccc}
	  Backbone &Method  &AP &AP$_{50}$  &AP$_{75}$ &AP$_{S}$  &AP$_{M}$  &AP$_{L}$ \\ \toprule
    \multirow{4}{*}{ResNet-18}
    &FP                           &33.2 &52.3 &34.8 &18.7 &35.6 &43.7 \\
   	&PACT~\cite{Choi2018PACTPC}   &25.3 &41.8 &26.0 &13.0 &26.8 &34.6 \\
    &APoT~\cite{Li2020AdditivePQ} &31.2 &50.1 &32.8 &{\bf 18.0} &33.5 &{\bf 40.6} \\
    &LCQ \small (Ours)            &{\bf 31.3} &{\bf 50.2} &{\bf 33.1} &17.6 &{\bf 33.8} &40.4 \\ \hline
    \multirow{3}{*}{ResNet-34}
    &FP                           &37.2 &57.0 &39.4 &21.4 &40.4 &48.9 \\
    &APoT~\cite{Li2020AdditivePQ} &35.2 &54.9 &37.1 &19.7 &{\bf 39.1} &{\bf 45.3} \\
    &LCQ \small (Ours)            &{\bf 35.5} &{\bf 55.3} &{\bf 37.6} &{\bf 20.5} &39.0 &45.0 \\ \hline
    \multirow{3}{*}{ResNet-50}
    &FP                           &38.3 &58.3 &40.9 &21.5 &42.4 &49.5 \\
    &APoT~\cite{Li2020AdditivePQ} &{\bf 36.1} &56.0 &{\bf 38.7} &21.2 &{\bf 40.4} &44.9 \\
    &LCQ \small (Ours)            &{\bf 36.1} &{\bf 56.2} &38.4 &{\bf 21.7} &39.9 &{\bf 46.1}\\ \hline
	\end{tabular}}
	\label{tab:ap_coco_3bit}
  \vspace{-0.8em}
\end{table}
\par Table~\ref{tab:ap_coco_3bit} shows the results of the 3-bit RetinaNet (W3/A3), where we observed that LCQ achieved comparable APs to those of the non-uniform quantization method, APoT~\cite{Li2020AdditivePQ} (our implementation).
\par These results show that LCQ can quantize the convolutional layers with less performance degradation even for the FPN architecture and heads connected to the regression and classification layers.
\begin{figure}[t]
  \setlength{\belowcaptionskip}{-1pt}
  \setlength{\abovecaptionskip}{-1pt}
  \centering
  \includegraphics[width=0.8\linewidth]{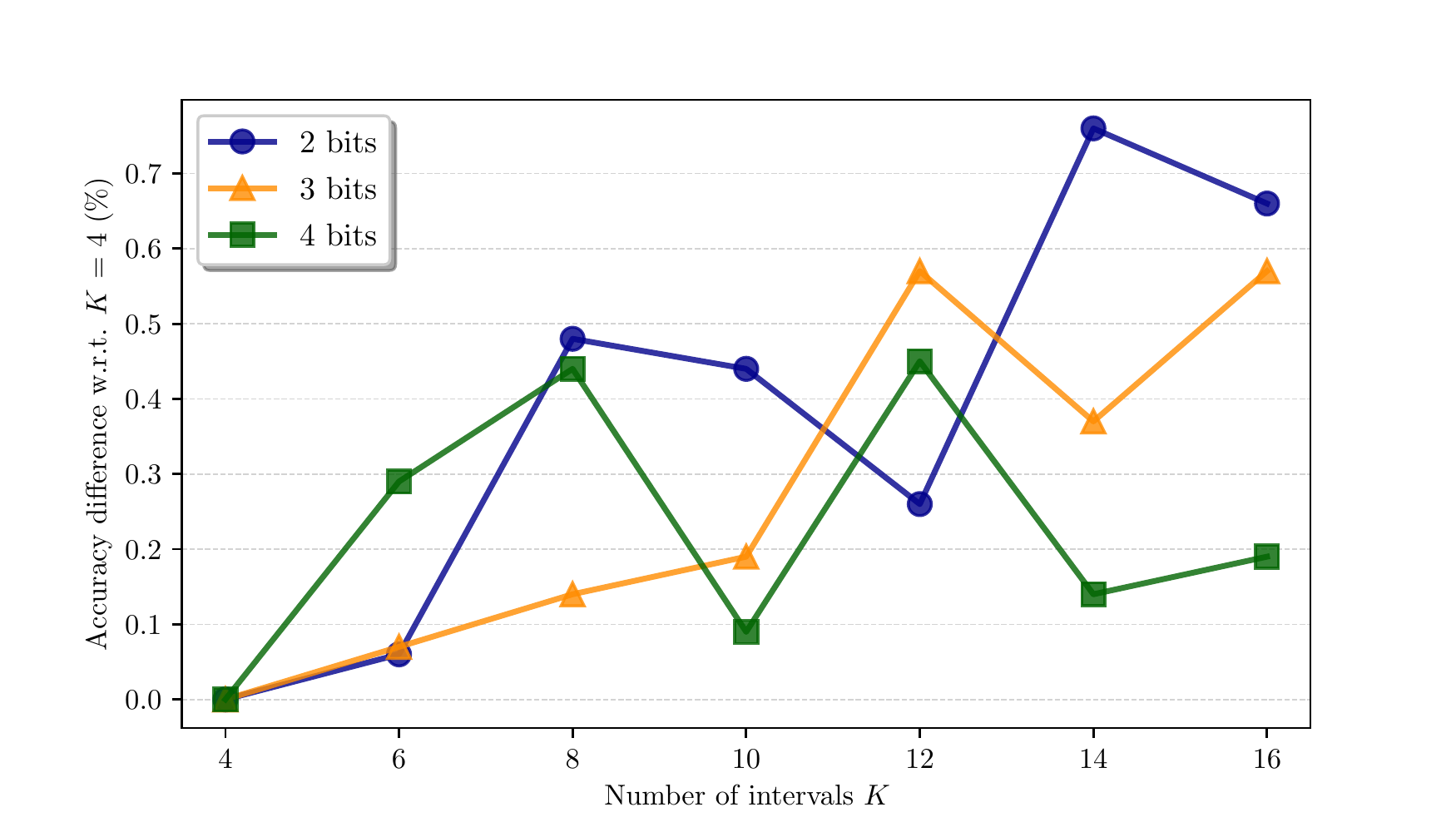}
  \caption{
  Relation between accuracy and number of intervals $K$ in LCQ for the ResNet-20 model on CIFAR-100.
  }
  \label{fig:ablation_intervals_gain}
  \vspace{-1em}
\end{figure}
\subsection{Ablation Studies} \label{sec:ablation}
{\bf Number of intervals.}
The piecewise linear function in the companding function requires the number of intervals $K$ as a predefined hyperparameter.
We tested the relation between number of intervals and prediction accuracy in the ResNet-20 model for the CIFAR-100 dataset under the same experimental conditions as in Sec.~\ref{sec:cifar10}.
Figure~\ref{fig:ablation_intervals_gain} shows the relative accuracy difference with respect to $K=4$ for each different number of bits when the number of intervals is increased from 4 to 16.
The accuracy with 2, 3, and 4 bits for $K=4$ was 65.2, 67.4, and 67.6, respectively.
For numbers of bits, accuracy tends to improve as the number of intervals increases, and we found that accuracy tends to significantly improve with fewer bits.
Since the companding function provides more flexibility in controlling quantization levels as the number of intervals increases, we infer that this flexibility is related to accuracy.
\begin{table}[hbt!]
	\centering
	\caption{Comparison of Top-1 accuracy (\%) w.r.t. LWN.}
	\scalebox{0.88}{
	\begin{tabular}{l|l|ccc}
		Model \& Data &Method &W2/A2 &W3/A3 &W4/A4 \\\toprule
		\multirow{2}{*}{\shortstack[l]{ResNet-20\\~on CIFAR-10}}
		  &\small LCQ w/o LWN     &91.4  &92.6  &93.1 \\
      &\small LCQ w/ LWN     &{\bf91.8}  &{\bf92.7}  &{\bf93.2} \\\hline
    \multirow{2}{*}{\shortstack[l]{ResNet-18\\~on ImageNet}}
      &\small LCQ w/o LWN    &68.6  &70.4  &{\bf71.5} \\
      &\small LCQ w/ LWN    &{\bf68.9}  &{\bf70.5}  &{\bf71.5} \\\hline
	\end{tabular}}
	\label{tab:ablation_lwn}
\end{table}
\par {\bf Effect of LWN.}
We used ResNet-18/20 and the CIFAR-10 and ImageNet datasets to investigate the effectiveness of LWN on accuracy. Table~\ref{tab:ablation_lwn} shows the results for different bit-widths.
Note that when LWN was not used, we instead applied the conventional method~\cite{Li2020AdditivePQ}, which only applies standardization to pre-quantized weights.
We observed that the results show a relatively large improvement at 2 bits and a minor improvement at 3- and 4-bits.
Thus, LWN is a simple yet reliable method that can contribute to accuracy.
\begin{table}[hbt!]
	\centering
	\caption{Comparison of the different outer bit-widths.}
	\scalebox{0.88}{
	\begin{tabular}{l|c|c|c}
		Model \& Data &$b'_w/b'_a$ &Acc. (\%) &LUT size (bytes) \\\toprule
		\multirow{3}{*}{\shortstack[l]{ResNet-20\\~on CIFAR-10\\($b_w=b_a=3$)}}
		  &8~/~8 &92.8  &42.0 \\
      &6~/~6 &92.8  &31.5 \\
      &4~/~4 &92.6  &21.0 \\\hline
    \multirow{3}{*}{\shortstack[l]{ResNet-18\\~on ImageNet\\($b_w=b_a=3$)}}
  	  &8~/~8 &70.6  &42.0 \\
      &6~/~6 &70.5  &31.5 \\
      &4~/~4 &70.4  &21.0 \\\hline
	\end{tabular}}
	\label{tab:ablation_outer_bit-widths}
	\vspace{-1em}
\end{table}
\par {\bf Outer bit-widths.}
Table~\ref{tab:ablation_outer_bit-widths} shows the top-1 accuracy and the memory cost of LUTs per layer for the 3-bit ResNet-18/20 model and the CIFAR-10 and ImageNet datasets for the outer bit-widths ($b'_w$ for weights and $b'_a$ for activations) described in Sec.~\ref{sec:lut}.
We observed that accuracy for CIFAR-10 remained the same from 8 to 6 bits, but degraded at 4 bits.
In contrast, accuracy for ImageNet showed a clearer decrease compared with that for CIFAR-10 and tended to decrease linearly as the number of outer bit-widths decreased.
However, there is still an advantage for both cases, especially at 4 bits, as accuracy degradation is as small as $0.2\%$ points and the memory cost of a LUT is half that of the 8-bit case.
\section{Conclusion} \label{sec:conclusion}
We proposed LCQ as a non-uniform quantization method that can optimize quantization levels via a learnable companding function.
We formulated the companding function so that quantization levels can be flexibly and non-uniformly controlled by training.
We also found that we can stabilize quantization training by limiting the effective scope of normalization to only the weight quantizer (LWN).
In addition, we reduced the memory cost of the LUTs required for the efficient inference by applying the re-quantization technique.
Various experiments involving image classification and object detection tasks for extremely low-bit models showed that LCQ achieved performance better than or comparable to conventional uniform and non-uniform quantization methods.
We also conducted three ablation studies.
The results showed that there is a likely proportional relationship between the number of intervals in the companding function and its accuracy, that LWN contributes to accuracy, and that accuracy can be maintained to some extent by reducing the outer bit-widths related to the LUT size.
While we showed that non-uniform quantization has strong potential, fast inference on resource-constrained devices requires an efficient hardware accelerator in practice, so we plan to tackle this problem in future works.
\section{Acknowledgement}
\label{sec:acknowledgement}
This paper is partly based on results obtained from a project, JPNP16007, commissioned by the New Energy and Industrial Technology Development Organization (NEDO).
{\small
\bibliographystyle{ieee_fullname}
\bibliography{refs}
}
\clearpage
\appendix
\twocolumn[
  \begin{@twocolumnfalse}
    \maketitle
    \begin{center}
    \textit{Supplementary Material for ``Learnable Companding Quantization for Accurate Low-bit Neural Networks''}
    \end{center}
  \end{@twocolumnfalse}
]
\setcounter{section}{0}
\setcounter{table}{0}
\renewcommand{\thesection}{\Alph{section}}
\renewcommand{\thetable}{S\arabic{table}}
\renewcommand{\thealgorithm}{S\arabic{algorithm}}
\section{Relations between LCQ and the conventional methods} \label{sec:supp_relation}
We clarify the difference between our proposed method and some similar conventional methods as follows:
LSQ~\cite{Esser2020LearnedSS} uses a learnable clipping function similar to our method, however, they use a uniform quantization function.
Therefore, the quantization levels are not learnable.
Since the input distribution of DNNs is usually not uniform, non-uniform quantization is better than uniform quantization to reduce the quantization error.
QIL~\cite{Jung2019LearningTQ} uses also the uniform quantization method that non-linearly transforms the input (corresponding to our ``compressing function") and then uniformly quantizes it, however, it does not apply the expanding function as we proposed.
Without the expanding function, the quantization error is likely to be large.
APoT~\cite{Li2020AdditivePQ} uses non-uniform quantization, however, their quantization levels are not learnable.
DQ~\cite{polino2018model} learns non-uniform quantization levels, however, their levels are optimized with simple heuristic gradients, while our levels are optimized with gradients based on the derivative of the companding function.
LQ-Nets~\cite{Zhang_2018_ECCV} also learns non-uniform quantization levels, however, it does not use gradients to optimize the levels, unlike our method.
\section{Training algorithm for LCQ} \label{sec:supp_algorithm}
When training quantized DNNs with LCQ, we independently apply the LCQ quantizer to the weights and activations for the convolutional or fully-connected layers.
Algorithm~\ref{alg:lcq} summarizes the LCQ training procedure for a convolutional layer as an example.
Note that ``$\ast$'' denotes a convolutional operation,
and that the LCQ parameters are given independently on a layer-by-layer basis.
\begin{algorithm}[hbt!]
  \caption{Training a convolutional layer with LCQ.}
  \label{alg:lcq}
  \begin{algorithmic}[1]
      \Statex {\bf Input:} full precision weights $w$ and full precision inputs/activations $a$, and the corresponding parameters: the clipping parameters $(\alpha_w, \alpha_a)$, the companding parameters $(\theta_w, \theta_a)$, the bit-widths $(b_w, b_a)$ and the outer bit-widths $(b'_w, b'_a)$.
      \Statex {\bf Output:} updated parameters $w$, $\alpha_w$, $\alpha_a$, $\theta_w$ and $\theta_a$.
      \State Compute the quantized weights using Eq.~(\ref{eq:lwn}) and Eq.~(\ref{eq:lcq_prime}): $w_q \leftarrow$ $\text{Quantize}(w, \alpha_w, \theta_w, b_w, b'_w)$.
      \State Compute the quantized activations using Eq.~(\ref{eq:lcq_prime}): $a_q \leftarrow \text{Quantize}(a, \alpha_a, \theta_a, b_a, b'_a)$.
      \State Compute the convolution outputs: $y \leftarrow w_q \ast a_q$.
      \State Compute the loss $\mathcal{L}$ and the gradients $\frac{\partial \mathcal{L}}{\partial y}$.
      \State Compute the gradients for the weights $\frac{\partial \mathcal{L}}{\partial y}\frac{\partial y}{\partial w}$.
      \State Compute the gradients for the clipping parameters $\frac{\partial \mathcal{L}}{\partial y}\frac{\partial y}{\partial \alpha_a}$ and $\frac{\partial \mathcal{L}}{\partial y}\frac{\partial y}{\partial \alpha_w}$ based on Eq.~(\ref{eq:deriv_clip}).
      \State Compute the gradients for the companding parameters $\frac{\partial \mathcal{L}}{\partial y}\frac{\partial y}{\partial \theta_a}$ and $\frac{\partial \mathcal{L}}{\partial y}\frac{\partial y}{\partial \theta_w}$ based on Eq.~(\ref{eq:chain_rule}) and Eq.~(\ref{eq:deriv_q_gamma}).
      \State Update $w$, $\alpha_a$, $\alpha_w$, $\theta_w$ and $\theta_a$ with the corresponding gradients, respectively.
  \end{algorithmic}
\end{algorithm}
\section{Validity of comparing LCQ and uniform quantization methods} \label{sec:supp_validity}
Our method assumes that multiplication is replaced by memory access to LUTs during inference, and the speed of the memory access depends on an efficient hardware accelerator design.
Therefore, with respect to the comparison between the proposed and conventional methods, it is difficult to evaluate theoretical metrics (\eg, FLOPs) for computational efficiency, and also to evaluate the actual speedup without dedicated hardware support.
However, since there is almost no difference between our method and conventional methods in terms of memory usage, we compare them in terms of accuracy at the same bit-widths.
This accuracy comparison is worthwhile because it allows us to evaluate the model's portability to memory-constrained devices.
\begin{table}[hbt!]
	\centering
	\caption{Comparison of memory usage with/without LUT in bytes. $b_w/b_a$ indicates the bit-width for weights and activations, respectively.}
	\scalebox{0.82}{
	\begin{tabular}{c|c|c|c|c}
		Model &$b_w/b_a$ &w/ LUT &w/o LUT &Diff. \\\toprule
    \multirow{3}{*}{\shortstack[c]{ResNet-18\\(44.59 MB in FP32)}}
      &2/2 &3.19 MB  &3.19 MB  &114 B \\
      &3/3 &4.52 MB  &4.52 MB  &798 B \\
      &4/4 &5.85 MB  &5.85 MB  &3990 B \\\hline
    \multirow{3}{*}{\shortstack[c]{ResNet-34\\(83.15 MB in FP32)}}
      &2/2 &5.63 MB  &5.63 MB  &210 B \\
      &3/3 &8.16 MB  &8.16 MB  &1470 B \\
      &4/4 &10.70 MB  &10.69 MB  &7350 B \\\hline
    \multirow{3}{*}{\shortstack[c]{ResNet-50\\(97.46 MB in FP32)}}
      &2/2 &7.73 MB  &7.73 MB  &312 B \\
      &3/3 &10.52 MB  &10.52 MB  &2184 B \\
      &4/4 &13.33 MB  &13.32 MB  &10920 B \\\hline
    \multirow{2}{*}{\shortstack[c]{MobileNet-V2\\(13.37 MB in FP32)}}
      &\multirow{2}{*}{4/4} &\multirow{2}{*}{2.41 MB}
      &\multirow{2}{*}{2.40 MB} &\multirow{2}{*}{10920 B}\\
      &&&& \\\hline
	\end{tabular}}
	\label{tab:model_sizes}
\end{table}
\par We then show the additional memory usage by LUTs is almost negligible.
For the models and bit-widths used in the experiments in this paper, Table \ref{tab:model_sizes} shows the memory usage of the LCQ models (w/ LUT) and the uniform quantization models (w/o LUT).
Note that we set 8 as the outer bit-width for both weights and activations.
Clearly, there is almost no difference in their memory usage for all the combinations of the models and the bit-widths.
\end{document}